\documentclass[preprint,12pt]{elsarticle}

\usepackage{multirow}
\usepackage{booktabs}
\usepackage[skip=4pt]{caption}   

\usepackage{amssymb}
\usepackage{amsmath}

\journal{Information Fusion}

\begin{document}

\begin{frontmatter}



\title{CausAdv: A Causal-based Framework for Detecting Adversarial Examples} 


\author{Hichem Debbi} 

\affiliation{organization={Laboratory of Informatics and applications of M'sila, Department of Computer Science, University of M'sila},
            addressline={PO Box 166}, 
            city={Ichebilia},
            postcode={28000}, 
            state={M'sila},
            country={Algeria}}

\begin{abstract}
Deep learning has led to tremendous success in computer vision, largely due to Convolutional Neural Networks (CNNs). However, CNNs have been shown to be vulnerable to crafted adversarial perturbations. This vulnerability of adversarial examples has motivated research into improving model robustness through adversarial detection and defense methods. In this paper, we address the adversarial robustness of CNNs through causal reasoning. We propose CausAdv: a causal framework for detecting adversarial examples based on counterfactual reasoning. CausAdv learns both causal and non-causal features of every input, and quantifies the counterfactual information (CI) of every filter of the last convolutional layer. We then perform a statistical analysis of the
filters’ CI across clean and adversarial samples, to demonstrate that adversarial examples exhibit different CI distributions compared to clean
samples. Our results show that causal reasoning enhances the process of adversarial detection without the need to train a separate detector. Moreover, we
illustrate the efficiency of causal explanations as a helpful detection tool by visualizing the extracted causal features. 
Code for reproducing our results is available at : https://github.com/HichemDebbi/CausAdv/tree/main.
\end{abstract}

%


\end{frontmatter}



\section{Introduction}
\label{sec:intro}

Deep neural networks (DNNs), which represent an major class of machine learning models have shown a remarkable success in many applications like image classification \cite{000}, natural language processing \cite{001}, and two-player games \cite{002}. Despite their high accuracy, they have a black box nature, raising serious concerns about extending their application to other critical domains like autonomous driving and avionics \cite{Tuncali} and healthcare \cite{Finlayson}. Unlike interpretable machine learning models like decision trees and Bayesian networks, understanding the inference process in DNNs is more challenging, due to the large number of learnable parameters, complex architecture, and various learning procedures. 

The black-box nature of DNNs has revealed one of their most critical vulnerabilities: adversarial examples.    \cite{Szegedy} and \cite{Biggio} show that every DNN model is vulnerable to adversarial examples, which refer to malicious perturbations on the input image. Although these perturbations are imperceptible to humans, yet they succeed to fool the state-of-the-art models \cite{Krizhevsky,Szegedy_Deeper_2015,Simonyan_2015,He_2016}. It has been found that adversarial examples can be applied across different models \cite{He_2016,Wu_2020,Wang_2021} and be harmful even in the real world \cite{Duan_Physic,Hussain_2024,Ali_Ghanem_2025}, which raises many safety concerns especially in autonomous driving \cite{Kevin_Physic} and healthcare \cite{Ma_Medical}.

The act of generating adversarial examples to threat the security of DNNs systems is called adversarial attacks. Adversarial attacks range in different categories.
Regarding adversary's knowledge, we have white and black box attacks. For white box attacks, the adversary has a full knowledge about the trained classification model including model architecture, hyperparameters, activation functions, and model weights. These attacks usually take the gradient of the predictions probabilities with respect to particular pixels. Some famous white box attacks are FGSM \cite{Goodfellow_FGSM}, PGD \cite{MandryPGD},
BIM \cite{KurakinBIM}, Square \cite{ACFH2020square} and C\&W \cite{Carlini_CW}. While most adversarial attacks are white-box attacks, adversarial examples can nevertheless be transferred to black-box settings due to their transferability property \cite{Papernot2016TransferabilityIM}. Several works have further improved this transferability by proposing more sophisticated attack strategies \cite{10858076}.


From the target label's perspective, we have targeted and untargeted attacks. Consider an input image $x$ that is fed into a classifier $M_{\theta}(.)$, where $\theta$ refers to the model's parameters. An adversarial example against  $M_{\theta}(.)$ is defined then as another image $\hat{x}$ such that $\parallel \hat{x} - x\parallel$ is small, but the classifier prediction is no longer the same, i.e $M_{\theta}(\hat{x})\neq M_{\theta}(x)$. While untargeted attacks aim to generate any $\hat{x}$ that fools the model, targeted attacks in contrast generate $\hat{x}$  given a specific prediction label $\hat{y}$ such that $M_{\theta}(\hat{x})=\hat{y}$ and $\hat{y} \neq M_{\theta}(x)$. The small difference between the original image $x$ and the adversarial one $\hat{x}$ is  subjected to a perceptibility threshold $\epsilon$ : $\parallel \hat{x} - x \parallel \leq \epsilon$, where $\epsilon > 0$. Here the perceptible difference $\| \cdot \|$ is usually obtained using the Euclidean norm  $\| \cdot \|_{2}$ or the max-norm  $\| \cdot \|_{\infty}$, which is reduced to an optimization problem. While $l_{\infty}$ measures the maximum change that can be made for all the pixels in the adversarial examples, $l_{2}$ norm measures the Euclidean distance between $x$ and $\hat{x}$. 

Many defense methods have been proposed in order to make CNNs more secure and robust against adversarial attacks. According to Papernot et al. \cite{Papernot2016TransferabilityIM}, the works on defending against adversarial examples can be grouped into two main categories: adversarial training and robust network architectures. 
In addition, we have detection-based methods, which aim to determine whether an input image adheres to certain criteria typically satisfied by natural images, even those that are misclassified, but they are violated by adversarially perturbed images. These methods address this issue from multiple perspectives: statistics\cite{Roth_Stat,Xin_Stat,Metzen,Feinman}, detector training \cite{MagNet,Nguyen} and prediction inconsistency\cite{Weilin_Squeez,Tao_Turning}.

Based on the assumption that causal reasoning could help to bring robustness, and thus it could cope with adversarial attacks, many works tried recently to investigate this direction \cite{Zhang_Causal,Tang_Causal}.  These works go beyond sample statistical correlations by revealing how perturbations exploit non-causal dependencies between features and labels. They are built on the assumption that aligning distributions against confounders enables models to focus on causally relevant features.

This paper proposes CausAdv: a causal-based framework for detecting adversarial examples in CNNs. Our approach consists of two main steps, in the first step we perform a causal learning process to identify causal robust features, then, in the second step, we perform statistical analysis based on the outcome of the first step. In the first step, we identify causally robust features by analyzing the contributions of convolutional filters to model predictions. Guided by the causality abstraction principle  \cite{Beckers28}, we focus on filters in the last convolutional layer as actual causes of classification decisions, since they refer to high level features. For each prediction, we classify filters as causal or non-causal and compute their Counterfactual Information (CI), which is a score reflecting the change in prediction probability when a filter is removed. This quantifies the filter's importance to the predicted class. In the second step, we design detection strategies based on statistical analysis of filters' CI distributions to discriminate between clean and adversarial inputs. Additionally, we demonstrate that visualizing causal features provides both explainability and enhances detection capability.


The main contributions of this paper are:
\begin{itemize}
	\item We propose CausAdv, a causal framework for detecting adversarial examples based on counterfactual reasoning. 
	\item CausAdv performs statistical analysis of Counterfactual Information (CI) to distinguish adversarial from natural samples.
	\item CausAdv is architecture-agnostic and training-free, requiring no modification to the model or input images.
	\item CausAdv achieves a high detection accuracy, especially under the BIM attack, providing also clear and interpretable causal explanations.
\end{itemize}

\section{Related work}

\textbf{Robustness-Based Defenses}

Although adversarial training (AT) remains the most effective strategy \cite{QIAN2022108889}, the internal activation behavior of DNNs under small perturbations is still not well understood, particularly how minor changes accumulate across layers to fool the network and how to be mitigated. To address this, several studies have proposed robust architectural designs \cite{Papernot_Distil,Gu_regul,Xu_activations_Cam}. One of the most used methods is \textit{gradient masking}, which aims to reduce sensitivity of DNNs to small perturbations. Papernot et al. \cite{Papernot_Distil} introduced the \textit{defensive distillation} methods, which involves training a large model, called the teacher, to maximize the likelihood of correct class predictions, and then transfer this knowledge to a smaller model, called the student. The student model is then expected to predict the same probability distribution over classes. So, relying on a hard softmax function could provide smooth decision boundaries, which makes it very hard to find adversarial examples without losing imperceptibility. While effective against attacks such as JSMA \cite{JSMA} on MNIST and CIFAR, this approach remains vulnerable to transferable black-box attacks. Gu and Rigazio \cite{Gu_regul} proposed \textit{gradient regularization} through contractive autoencoder trained on adversarial examples prior to classification. This method imposes a layer-wise contractive penalty to constrain the model's sensitivity. However, the effectiveness of this approach depends on the class of adversarial examples used during training, and in some cases, the combined autoencoder-classifier model may fail.

Other works tried to build robust architectures through different techniques such as batch normalization \cite{Galloway_Batch} and skip connection \cite{Wu_Skip}. Recent work has also investigated defense against adversarial attacks from the activations perspective. These include different activation functions \cite{Zhang_activations} and new activation operations \cite{Wang_activations}. Xu et al. \cite{Xu_activations_Cam} explored through interpretation the promotion, suppression and balance effects adversarial perturbations on neurons' activations, and recently Bai et al. \cite{Bai_activations} by proposing Channel-wise Activation Suppressing (CAS) to mitigate redundant activations exploitable by attacks.

Beyond modifying activations, several methods focus on processing activation outputs. Stochastic Activation Pruning (SAP) \cite{Dhillon_Stochactivations} takes each activation with a probability proportional to its absolute value. Xiao et al. \cite{Xiao_activations} proposed a new activation function k-Winner-Takes-All (kWTA) that takes only top-K outputs in every activation layer. Adversarial Neural Pruning (ANP) \cite{Madaan_activations} tries to remove activation outputs that are vulnerable to adversarial examples using Bayesian method. Feature Denoising (FD) \cite{Xie_2019_CVPR} alters the CNN architecture by adding new blocks, with the aim of denoising the feature maps. However the modified architecture needs to be retrained on adversarially generated samples. Similar pixel-level denoising approach was earlier proposed by Fangzhou et al. \cite{Fangzhou_Reg}. 


\textbf{Consistency-Based Defenses}

Feature squeezing by Xu et al. \cite{Weilin_Squeez} argues that input feature spaces are excessively large, providing adversaries with greater opportunities to manipulate inputs in high-dimensional spaces. To counter this, the authors propose reducing redundant input features through transformations such as non-local means and median smoothing, producing semantically equivalent variants of the original image. The model's predictions are then compared across these variants, If the model's prediction on the original input differs significantly from its predictions on the transformed versions beyond a predefined threshold, the input is flagged as adversarial. Additional squeezing techniques include reducing the image's color bit depth from 8-bit to 1-bit, which retains visual similarity while potentially eliminating adversarial noise.


Yu et al. \cite{Tao_Turning} suggest that, by inheritance, valid adversarial perturbations exist around natural images and can be used as signatures to indicate whether an input image is adversarial. They identify two main properties of natural images: first, robustness to random noise, which can be tested by adding Gaussian noise and observing the prediction probabilities; second, proximity to the decision boundary of their class, which can be assessed by counting the number of gradient steps required to induce a class change. These two properties form the basis for their detection strategy, which flags adversarial examples when observed deviations exceed predefined thresholds. Although the authors demonstrate that adversarial examples can be optimized to appear more natural and evade detection with sufficient gradient queries, their method shows promising results against white-box attacks. However, it suffers from a high false positive rate, and the second property requires extensive computation, limiting its practical applicability. This approach shares similarities with \cite{Weilin_Squeez}, as both modify the input image. In contrast, our method avoids altering the input, modifying the architecture, and retraining. Instead, we rely solely on a causal learning process as a first step.

\textbf{Statistical-Based Detection}

Roth et al. \cite{Roth_Stat} proposed a statistical-based strategy for detecting adversarial examples by analyzing logit stability under noise perturbations.  For an input image $x$, the method tries to compare its logit vector $f_{z}(x)$ to its noisy logit vector $f_{z}(x + \eta )$  where $\eta \sim \mathbb{N}$ is a noise added to the input from some distribution $\mathbb{N}$ such as Bernoulli and Gaussian noise. The core idea is that natural images exhibit stable logit responses under noise ($f_{z}(x) \approx f_{z}(x + \eta)$), whereas adversarial inputs cause significant deviations. Detection is performed using a threshold tuned to maximize the true positive rate. While this gradient-based approach the idea of statistical analysis with ours, it fundamentally differs: their method requires input-space perturbations and relies on gradient stability (logit vectors), whereas ours analyzes robust causal features through measuring filter contributions (CI). Moreover, their threshold selection is empirically chosen, whereas our statistical tests are grounded in causal theory. We argue that causal reasoning offers greater robustness, as it captures model-internal decision logic rather than input-output correlations.


Li and Li \cite{Xin_Stat} proposed a fully statistical approach for adversarial detection based on Principal Component Analysis (PCA) of convolutional features. For each convolutional layer, PCA is applied to the extracted feature maps, yielding a K-dimensional vector, where K corresponds to the number of filters. Statistically, they have shown that given the VGG architecture, more than 80\% of natural samples can be identified using the first convolutional layer. However, to do so, they need to build and train a \textit{cascade} classifier, which acts on the features extracted from every convolutional layer, and then concatenate and pass to the next convolutional layer. Both the input and convolved feature maps are treated as pixel-value distributions, under the assumption that natural samples exhibit Gaussian-like projections, whereas adversarial examples deviate significantly from this distribution. Detection thus relies on identifying these statistical deviations across PCA representations learned from natural and adversarial samples.

Compared to our approach, their method exploits all intermediate layers, which are trained jointly through a cascade classifier. In contrast, our method focuses exclusively on the last convolutional layer, a design choice that is motivated by the causal abstraction perspective. Unlike their model, CausAdv operates in a plug-and-play manner without training an additional detector, and relies only on natural samples for causal learning, making it more lightweight and adaptable. Moreover, their evaluation is limited to a single adversarial attack, the L-BFGS algorithm \cite{Szegedy}, without investigating widely used white-box attacks like PGD, FGSM, and BIM. Given its reliance on transformation techniques, their method shares conceptual similarities with previously discussed transformation-based defenses that require modifying the input image.

\textbf{Causal-Based Defenses}

While causal reasoning has been widely adopted recently into deep learning for various tasks such as explanation \cite{Narendra19,CXPlain,Debbi}, its application to adversarial robustness remains limited. Zhang et al. \cite{Zhang_Causal} address this issue from a causal perspective by assuming that adversarial examples result from a specific type of distributional shift in natural data. Therefore, they propose an adversarial distribution alignment method through building a causal graph representing the data generation process. The graph includes natural and perturbed inputs, causes grouped into two categories (content information (C), and style information (content-independent)(S)), the label (Y), the intervention (E), and finally the network's parameters ($\theta$). 

Each intervention $E$ includes a different data distribution, natural or adversarial, where spurious correlation between $Y$ and $S$ are is key indicator of adversarials. To mitigate this, the authors propose an adversarial distribution alignment method that detects and reduces such spurious dependencies, representing style information with Gaussian distributions as in \cite{Gal,Kendall}. Their work primarily analyzes the adversarial generation process and the role of spurious correlations rather than eliminating these vulnerabilities, aligning with our focus on detection. In contrast, our approach examines the distributions of filter Counterfactual Information (CI) to distinguish between natural and adversarial samples, explicitly incorporating counterfactual reasoning. 


Moreover, their framework requires training, making it a variant of adversarial training. In contrast, our method operates in a plug-and-play manner and does not require adversarial data for training. Their approach also relies on identifying spurious correlations between labels and style information $S$, which may not be effective when $S$ is not well represented, as is the case with simpler datasets like MNIST. Additionally, their use of a causal graph introduces scalability challenges, especially when dealing with high-dimensional data, which limits practical applicability. This is reflected in their experimental scope, which is restricted to CIFAR, with no evaluation on larger and more complex datasets such as ImageNet, which is actually a common limitation observed in many causality-based approaches.

Moreover, Zhang et al.’s method requires training, making it a variant of adversarial training, whereas our framework is plug-and-play and relies solely on natural samples. Their reliance on style features limits performance on datasets where such features are sparse (e.g., MNIST), and their causal graph design scales poorly to high-dimensional data. Consequently, their evaluation has been performed only on CIFAR, representing a limitation of current causality-based defenses that have yet to demonstrate scalability to large datasets such as ImageNet.

Building on the same assumption that the vulnerability to adversarial attacks stems from spurious correlations or confounding effects, Tang et al. \cite{Tang_Causal} also leverage causal reasoning. While Zhang et al. \cite{Zhang_Causal} leave the elimination of spurious correlations unaddressed, Tang et al. \cite{Tang_Causal} aim to actively suppress the learning of these confounders using a causal regularization technique called CiiV. This approach encourages the model to learn causal features by penalizing reliance on confounding variables.
In their framework, confounders are treated as potential causes and incorporated into a causal graph. As is typical in causal frameworks, estimating causal effects requires interventions. To this end, they rely on the instrumental variable estimation as a means of intervention. Both the confounder $C$ and the instrumental variable $R$ are embedded into the input image. The CiiV model must be trained, and its loss is combined with the standard cross-entropy loss, yielding improved adversarial robustness as demonstrated on CIFAR-10.


\section{Causality for detecting adversarial Examples}

\subsection{Introducing causality into CNNs} 

It is well known that the most important layers of CNNs are the convolution layers, which include filters. Filters represent the basic elements of the network,  which are crucial for generating activations in response to different regions of input images. Researchers have explored various aspects of CNNs to understand the effects of filters. In our framework we also investigate filters by considering them as the actual causes. 

\begin{figure}[h!]
	\centering
	\includegraphics[width=\linewidth]{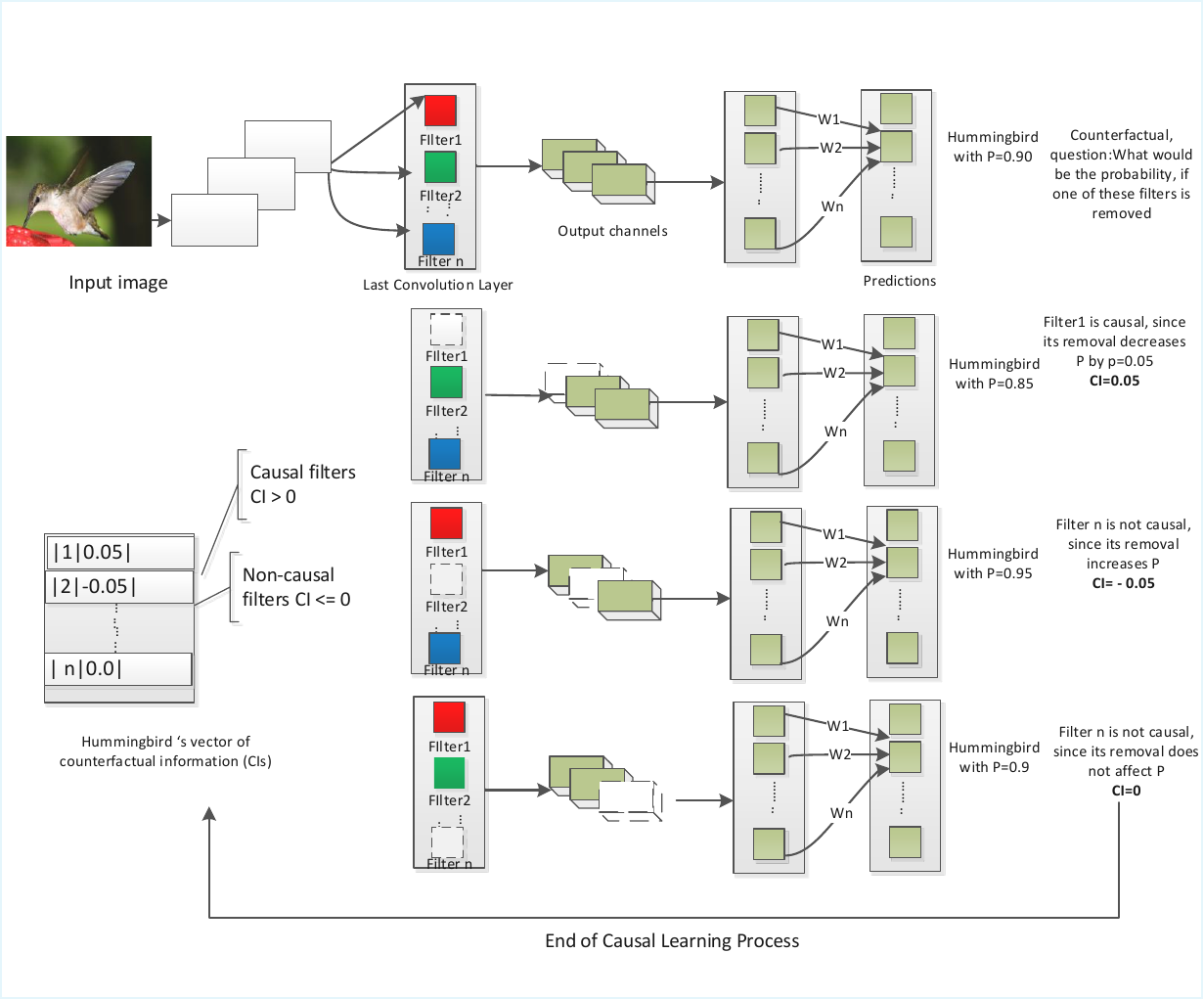}
	\caption{Causal learning process: measures the impact of each filter of the last convolutional layer of the CNN architecture when removed on the prediction probability. The difference in prediction probabilities after filter removal is referred to as counterfactual information (CI). Filters are categorized as causal or non-causal based on their effect on prediction probabilities.}
	\label{fig:Architecture}
\end{figure}


Halpern and Pearl have extended the definition of counterfactuals to build a rigorous model of causation, which is called a structural equations \cite{Halpern23}.
This definition of causality can be adapted to any CNN architecture by considering filters as actual causes for a predicted $\varphi$. However, since a class is  predicted with a probability $P$, $\varphi$ should not be just a Boolean formula that refers to the class predicted. Therefore, we should define countefactuals in a probabilistic setting. To do so, we should ask the following question: what would be the prediction probability, when one of the filters $f$ is removed at a specific layer $l$, while keeping all the filters intact ?

The initial step in our architecture is to associate each class with a prototype image. This prototype image is automatically selected by the CNN architecture as the instance with the highest confidence score $P$ among all available instances. The underlying intuition is that such a prototype captures the most salient and representative features of its class, making it easier for the CNN to recognize and classify it with high accuracy.


Now we can introduce the definition of an actual cause in CNNs.

\textbf{Actual cause}.
Let $F$ be the set of all filters, and $f_l\in F$ be a filter at a specific layer $l$. $f_l$ can be considered causal for a prediction $\varphi$, if its own removal decreases the prediction probability $P$ to $P'$, while the rest of filters are not modified. We consider the difference $P-P'$ as the counterfactual information, and it is denoted by $CI$. A filter is causal if its $CI>0$ and it is not causal otherwise(See Figure ~\ref{fig:Architecture}).

Based on the definition of causality \cite{Halpern23}, the set of variables can be partitioned into two sets, causal variables $W$, which have actual contribution to the Boolean formula $\varphi$, and non-causal variables $Z$. Accordingly, in our setting, the set of filters $F$ can be partitioned into two sets $F_{W}^{\varphi}$ and $F_{Z}^{\varphi}$, where the set $F_{W}^{\varphi}$ refers to filters causing the prediction $\varphi$, i.e. their removal will decrease the prediction probability $P$, these refer to causal features, whereas $F_{Z}^{\varphi}$ are not causal, in way that the removal of a filter in $F_{Z}^{\varphi}$ does not affect $P$, or otherwise increases it, and these refer to non-causal features. When the removal of a filter leads to increasing $P$, this means that this filter has a negative effect on the decision, which means that it is responsible for increasing the prediction probability of another class, not the current class. 
In the following, we will refer to $F_{W}^{x\mapsto \varphi}$ and $F_{Z}^{x\mapsto\varphi}$ as the causal and non-causal features of an input image $x$
predicted as $\varphi$ by a CNN architecture.

In CNN architectures, while early layers capture edges and textures, the last layers progressively learn complex features, since their filters are applied after many convolutions steps on the input image, making them well-suited for capturing high-level semantic information. Several techniques like CAM and Grad-CAM \cite{Selvaraju08} employ commonly the last convolutional layer to generate interpretable heatmaps, highlighting the most discriminative regions of the input image. In other words, the last convolution layer outputs the high level features, since it lays in the top hierarchy of the causal network. According to Beckers and Halpern \cite{Beckers28}, causal models exhibit an abstraction structure, where micro-level variables causally affect macro-level variables. This notion of abstraction plays a key role in our framework. Specifically, we focus solely on the filters of the last convolutional layer, treating them as macro causes within our causal reasoning process.


\subsection{Detection Strategies}

Many works such that \cite{Ilyas_Bugs} stated that classifiers try to use any possible activations in order to maximize distributional accuracy, even those activations related to incomprehensible features to humans. As a result, the adversarial attacks try to exploit the existence of the non-robust features learned by different architectures. Consequently, many existing defenses focus on modifying the input images to suppress or remove such non-robust features in order to improve robustness against adversarial attacks.

In contrast, CausAdv addresses this challenge through a causal learning framework, which emphasizes the extraction of causally robust features, those that are more stable under distributional shifts and adversarial perturbations. Our results demonstrate that these features are instrumental in distinguishing natural from adversarial examples. Furthermore, we show that statistical analysis of the Counterfactual Information (CI) distributions for clean and adversarial samples offers meaningful and discriminative patterns for detection.

Importantly, our approach relies solely on statistically analyzing the CI vectors, and employs a range of strategies to detect adversarial instances, without any modification of the input images.

After defining causal and non-causal filters, which represent the main concepts of our causal view on robustness, in the following, we define four main strategies for detecting whether an input image $x$ is an adversarial or it is natural, based on analyzing its causal ($F_{W}^{x\mapsto \varphi}$) and and non-causal filters ($F_{Z}^{x\mapsto\varphi}$).

\textbf{Strategy 1: Causal Features Existence}:
We consider an input image $x$ predicted as $\varphi$ adversarial if $F_{W}^{x\mapsto \varphi}=\emptyset$, or instead: $|F_{W}^{x\mapsto \varphi}|<=n$, where $n$ is a small natural number, which means that $x$ has no causal features, or at most has a very few number of causal features, or in other words all the CI are equal or less than zero. 

The core idea is that it is not possible to find a sample entirely devoid of causal features. This has been validated experimentally by computing the Counterfactual Information (CI) vectors for every class among the 1,000 prototype (clean) classes of the ImageNet dataset, as well as for the 10 classes of CIFAR10.
This applies even for classes with small number of features, which could happen in some basic objects having a low number of features, or objects having low CI such as the ones having fine-grained features such as flowers. Such a detection strategy would be very efficient as we will see later in the experiments section particularly with the BIM attack.

\textbf{Strategy 2: Correlation analysis (Pearson's coefficient)}:

Let us denote by $x_{\varphi}^{prot}$ the prototype image that corresponds to a class $\varphi$, a representative of inputs classified as $\varphi$. Given an input image $x$, we compute its Counterfactual Information (CI) distribution, denoted $c_{x}$, which encodes the contribution of each filter. We then define $x$ to be adversarial if its CI distribution is not sufficiently correlated with that of its class prototype $c_{x_{\varphi}^{prot}}$ with respect to a threshold $\tau$. Formally: $x$ is natural if $\rho (c_{x}, c_{x_{\varphi}^{prot}} )\geqslant \tau$ and adversarial otherwise, where $\rho(\cdot,\cdot)$ is the Pearson correlation coefficient.


$x_{\varphi}^{prot}$ is identified as prototype for the class $\varphi$ from the dataset among the instances of $\varphi$, as the one having the high prediction confidence according to the same CNN architecture. After conducting the causal learning process, we will have all causal and non-causal filters of each class, and these refer to related and non-related features to this class respectively. To assess the similarity between an input image and its prototype, we compare their CI vectors using a correlation-based metric, such as the Pearson correlation coefficient. This quantifies the degree to which the features (as captured by the CI vectors) are aligned. A low correlation implies that the input may contain non-causal or spurious features, indicating a potential adversarial perturbation.

This idea bears some resemblance to statistical approaches such as that of Grosse et al. \cite{Grosse}, who proposed measuring statistical distances between large sets of adversarial and legitimate inputs to detect individual adversarial examples. However, in contrast to their method, which relies on aggregating statistics over entire datasets of legitimate inputs, our strategy considers only a single legitimate prototype image per class. Moreover, our analysis is based on causal features, which represent high-level, robust abstractions, rather than relying on gradient-based information, which is often noisy and less stable under adversarial perturbations. This distinction makes CausAdv inherently more robust and significantly improves its detection performance, as will be shown in our experimental results.

\textbf{Strategy 3: Zero effect}:
Let $F_{Z_{0}}^{x\mapsto\varphi}$ be the set of zero filters of an input image $x$ (filters whose CI=0) and a natural threshold $n$. $x$ is decided to be adversarial if its number of its zero filters $|F_{Z_{0}}^{x\mapsto\varphi}|< n$. 

This strategy is based on the observation that an input image may be adversarial if it lacks zero-CI filters, implying that the adversary is exploiting non-causal features to influence the prediction. Similar findings have been reported in previous works, where certain channels were shown to be over-activated in the presence of adversarial inputs.


To determine whether an input is suspicious, we introduce a threshold $n$, which represents the minimum expected number of zero-CI filters. This threshold is estimated by analyzing the behavior of the prototype image $x_{\varphi}^{prot}$.If the number of zero-CI filters in the input is significantly lower than that of the prototype, it may indicate adversarial manipulation. This is because, under normal conditions, no class exhibits an entirely positive or entirely negative CI vector, rather, each class has a typical range of zero-CI filters—those that neither contribute to nor detract from the class prediction.

This observation is further validated through experiments on CIFAR-10, where the reduction in zero-CI filters reliably correlates with adversarial inputs. For ImageNet, while the features are more abstract and high-level, we still consistently observe an acceptable number of zero-CI filters, supporting the robustness of this strategy across datasets.

\textbf{Strategy 4: Common Robust Causal Features}:
Let $x$ be an input image and $F_{W}^{x\mapsto\varphi}$ its set of causal features. We denote by $F_{W_{n}}^{x\mapsto\varphi}$ the top $n$ causal features of $x$, and $F_{W_{m}}^{x_{\varphi}^{prot}\mapsto\varphi}$ the top $m$ causal features of the $\varphi$'s prototype $x_{\varphi}^{prot}$. $x$ is said to be adversarial example if $F_{W_{n}}^{x\mapsto\varphi} \cap F_{W_{m}}^{x_{\varphi}^{prot}\mapsto\varphi} = \emptyset$, such that $m>=n$. 

That is, an input image $x$ considered adversarial if it fails to contains at least the top $n$ causal filters that are present in the top $m$ causal filters of the prototype image $x_{\varphi}^{prot}$, where $m>=n$. In our experiments, we explore several settings for $n$ and $m$ such as $(3,10)$, $(5,20)$ and $(10,30)$ respectively. Across all these settings, we constantly observe a clear distinction between clear and adversarial samples. The underlying intuition is that the top causal features represent the most robust features that should be present in any instance of the class.  Their absence may indicate that an adversary is attempting to increase the prediction confidence for an incorrect class by leveraging non-causal or spurious features.

Our strategy for analyzing causal and non-causal features shares conceptual similarities with the work of Bai et al. \cite{Bai_activations}, who detect adversarial inputs by identifying anomalous activations across all layers of a CNN. Their method enhances robustness by removing or promoting certain channels based on their reliability. However, unlike their architecture-wide analysis, our approach is more targeted and lightweight, focusing exclusively on the last convolutional layer in accordance with the causal abstraction principle. This allows us to isolate high-level robust features without modifying the architecture or requiring complex channel tracking across layers.

\section{Experiments}

\subsection{Experiment Setup}

\textbf{Data and Attacks:}
We conduct our experiments on ImageNet\cite{000} and CIFAR-10 \cite{CIFAR}.
ImageNet consists of 1000 diverse classes. We sample 6 random images for all the 1k classes within the ImageNet validation dataset, resulting in 6000 samples. For CIFAR, we consider the entire test set, which consists of 10,000 samples.

\textbf{Architecture:}

We perform our experiments on ImageNet using the VGG16 architecture. Specifically, we focus on the last convolutional layer, denoted as conv13, to compute the Counterfactual Information (CI) of the filters. CI values are computed for all 1k prototype images in ImageNet and the 10 prototypes in CIFAR-10, as well as for all 6000 evaluation samples from ImageNet and 10,000 samples from CIFAR. In the case of adversarial examples, CI values are computed separately for each attack type, allowing for a detailed comparative analysis between clean and adversarial instances.

\textbf{Attacks Setting:}

We evaluate CausAdv on both ImageNet and CIFAR-10 against both targeted and untargeted variants of four widely used $l\infty$ bounded adversarial attacks: Fast Gradient Sign Method (FGSM)\cite{Goodfellow_FGSM}, Projected Gradient Descent (PGD)\cite{MandryPGD}, Basic Iterative Method (BIM)\cite{KurakinBIM}, C\&W \cite{Carlini_CW} and Square Attack \cite{ACFH2020square}. While the first four methods can be configured to be both targeted and untargeted, Square attack is originally a recent untargeted attack. For ImageNet targeted attacks, we randomly assign unique target labels, one for each of the 1k selected source classes, ensuring that no target label overlaps with its corresponding source class. A similar procedure is used for CIFAR-10, where each of the 10,000 samples is assigned a distinct target class (different from its true label), covering all 10 classes.

To ensure the attacks truly change the model's prediction, we carefully select the perturbation budget $\epsilon$. For ImageNet, we found that $\epsilon = 4$ was insufficient to reliably induce misclassification across all attacks. Therefore, we use $\epsilon=8$, which achieved an acceptable attack success rate (ASR) for both targeted and untargeted variants across all attacks. While FGSM, PGD, BIM and Square attack share a similar configuration based on the perturbation budget $\epsilon$, C\&W differs in that it is an optimization-based attack and is therefore configured via the number of optimization iterations, which is set to 500. Under these settings, FGSM, PGD, and BIM achieve high ASR, whereas Square Attack and C\&W attain lower ASR. 

For CIFAR-10, due to its higher resilience to adversarial perturbations compared to ImageNet \cite{Ghorbani_Abid_Zou_2019}, we set a larger perturbation budget of  $\epsilon=24$ to achieve a meaningful attack rate. Despite this higher value, we did not really reach a high ASR, highlighting CIFAR's relative robustness under these settings.

All experiments are implemented using the Keras framework, and adversarial attacks are generated using the Adversarial Robustness Toolbox (ART) library \cite{ART}, which supports a wide range of attack algorithms and configurations.

\subsection{Experimental results}
\begin{figure}
	\centering
	\includegraphics[width=\linewidth]{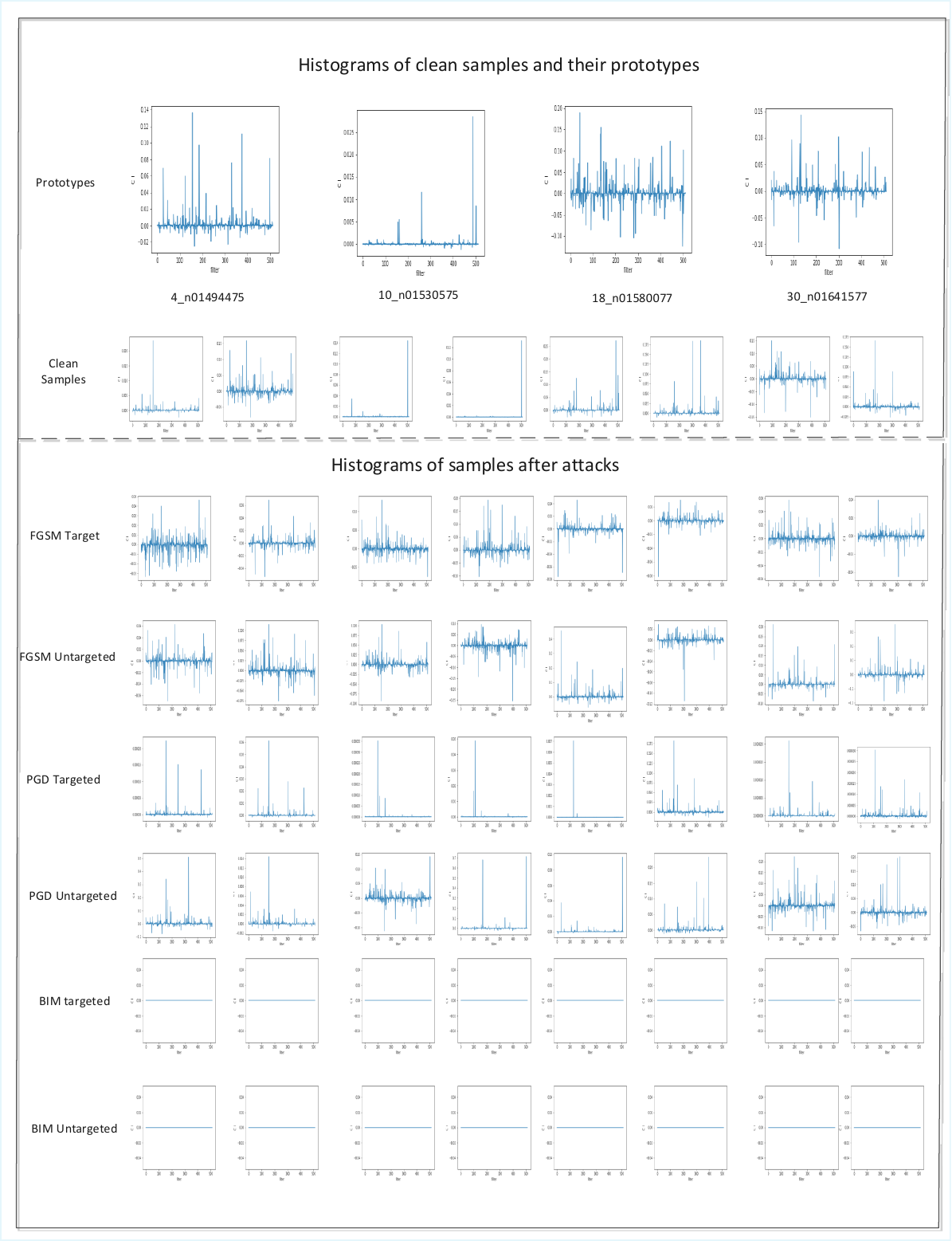}
	\caption{Histograms of CI distributions for some ImageNet samples with their prototypes (2 samples per class) and their adversarial examples. All the attacks are performed with $\epsilon=8$.}
	\label{fig:Histograms}
\end{figure}

We perform the causal learning process and compute the CI vectors for all prototypes, clean samples, and adversarial samples. Histogram plots of selected samples are shown in Figure~\ref{fig:Histograms}. The x-axis indicates to the number of filters, while the y-axis represents the corresponding CI values.

Initial analysis of the results reveals that clean samples generally exhibit approximately normal distributions, closely resembling those of their corresponding prototypes. Regarding adversarial attacks, both targeted and untargeted BIM attacks are readily detected, as they fail to preserve causal features—a distinction effectively captured by \textit{Strategy 1}. The only attacks that manage to bypass detection without significantly altering causal features are PGD attacks, whose CI distributions remain similar to those of the prototypes. In contrast, FGSM attacks display noticeable distortions and anomalous behavior compared to both clean samples and prototypes. Specifically, FGSM tends to flip the causal status of many features, rendering several causal features non-causal and vice versa.

Based on the CI vectors of all samples, in the following we will present the experiments details with respect to every detection strategy.

\textbf{Strategy 1: Causal Features Existence:}

Using this strategy, we observe that adversarial samples are detected across all types of attacks. In particular, the BIM attack is characterized by a complete absence of causal features at $\epsilon = 8$, resulting in a 100\% detection rate for both targeted and untargeted variants on ImageNet. Other attacks, while still detectable, exhibit comparatively lower detection rates. Even when reducing the perturbation budget to $\epsilon = 4$, the results remain consistent. The BIM attack continues to exhibit a complete absence of causal features, leading also to a 100\% detection rate, while it maintains a high attack success rate.

We try to increase $\epsilon=8$ to $\epsilon=16$, the results for targeted and untargeted PGD attacks become similar to BIM, where all filters have nearly Zero CI values. Given this strategy, the detection rate will be also $100\%$ for PGD attack. It is worth noting that the best detection results for the PGD attack on ImageNet given max perturbation $\epsilon = 16$ are achieved by the ResNet-152 Denoise model\cite{Xie_2019_CVPR} with detection rate $42.80$. 


As noted in \cite{Dong_Benchmark}, PGD and BIM are closely related and often demonstrate comparable performance. Our findings under this detection strategy are consistent with this observation: both PGD and BIM are more effective than FGSM in generating adversarial samples. However, they are also more easily detected through causal feature analysis. Specifically, BIM is detected even at low perturbation budgets, whereas PGD exhibits greater robustness against this detection strategy. Notably, PGD only becomes as easily detectable as BIM when the perturbation budget is increased from $\epsilon = 8$ to $\epsilon = 16$.


\textbf{Strategy 2: Correlation analysis (Pearson coefficient)}

We test first the efficiency of this technique as a similarity technique for natural images, for both the CIFAR-10 and ImageNet datasets. 
For ImageNet, as we are dealing with $1000$ classes rather than 10 classes, we measure correlation with the top-5 classes, clean images and their corresponding prototypes on one hand, and between the CI vectors of adversarial samples and their prototypes on the other. The results are presented in Table~\ref{tab:PearsonImag}. While we report in the second column the attack success rate (ASR), in the third column, we report the number of samples whose top-5 most correlated prototypes include the predicted label $\varphi$. A sample $x$ is considered suspicious if its predicted class $\varphi$ does not appear among the top-5 correlated prototypes.  


\begin{table}[t]
	\renewcommand{\arraystretch}{1.0}
	\centering
	\begin{tabular}{lcc}
		\toprule
		\textbf{Samples} & \textbf{Attack Success Rate (ASR)} & \textbf{Well-correlated} \\
		\midrule
		Clean &  & 4,211 \\
		\midrule
		\multicolumn{3}{c}{\textit{Targeted}} \\
		\midrule
		FGSM & 5483/6,000 & 514 \\
		PGD & 5558/6,000 & 855 \\
		BIM & 5589/6,000 & 0 \\
		C\&W & 1998/6000 & 2032 \\
		\midrule
		\multicolumn{3}{c}{\textit{Untargeted}} \\
		\midrule
		FGSM & 5,613/6,000 & 1,099 \\
		PGD & 5,371/6,000 & 1,032 \\
		BIM & 5,784/6,000 & 0 \\
		C\&W & 2174/6000 & 1902 \\
		Square & 3,451/6,000 & 2,041 \\
		\bottomrule
	\end{tabular}
	\caption{Pearson-correlation-based results on 6,000 ImageNet samples (6 images per class), considering the top-5 predicted classes.}
	
	\label{tab:PearsonImag}
\end{table}

As shown in the results, $4,211$ of $6,000$ clean samples have their true class among the top-5 most correlated prototypes, which proves the usefulness of the Pearson correlation measure over CI distributions.  In contrast, adversarial samples generally exhibit lower top-5 correlation counts except in the case of C\&W and Square attack, which yield the highest number of matching samples at $2032$ and $2041$ respectively. However, it is worth noting that given the perturbation budget $\epsilon=8$, C\&W and Square attack do not attend a high ASR comparing to other methods, and this justifies the high number of top-5 correlated labels. To ensure a fair comparison, it is important to note that the $6,000$ samples were randomly selected.

For both targeted and untargeted BIM attacks, detection is straightforward, as the corresponding CI vectors are entirely null, making these adversarial samples easily identifiable. The strongest result is observed for the targeted FGSM attack, which achieves a higher detection rate with only $514$ top-5 predicted classes have been found. This behavior can be explained by the nature of the FGSM targeted attack: it tends to exploit any available activation in an isolated and unstructured manner to force the decision boundary closer to the target label, even if the prediction is of low confidence. Consequently, these perturbed features lose causal consistency, resulting in CI vectors with no meaningful alignment to any class prototype.

For the CIFAR-10 dataset, the results are reported in Table~\ref{tab:PearsonCIFAR}. While ASR is reported in the second column, the third column reports the number of samples with a Pearson correlation coefficient $\rho \geq 0.5$ between the sample and its class prototype. Unlike ImageNet, the top-5 correlation measure is not applicable here due to the limited number of classes (10). As shown in the results, clean samples exhibit the highest correlation count ($8,986$ out of $10,000$), which is significantly higher than any adversarial setting. The closest is the untargeted PGD attack, with $3,026$ samples exhibiting acceptable correlation with their prototype class. These results highlight the effectiveness of the similarity-based measure, especially in the CIFAR-10 setting, where the separation between clean and adversarial samples is more pronounced across all attacks.  

\begin{table}[t]
	\renewcommand{\arraystretch}{1.0}
	\centering
	\begin{tabular}{lcc}
		\toprule
		\textbf{Samples} & \textbf{ASR} & \textbf{Well-correlated} \\
		\midrule
		Clean &  & 8,986 \\
		\midrule
		\multicolumn{3}{c}{\textit{Target}} \\
		\midrule
		FGSM & 983/10,000  & 1,544 \\ 
		PGD & 709/10,000 & 1,372 \\
		BIM & 609/10,000 & 1,157 \\
		\midrule
		\multicolumn{3}{c}{\textit{Untargeted}} \\
		\midrule
		FGSM & 6,618/10,000 & 1,264 \\
		PGD & 1,049/10,000 & 3,026 \\
		BIM & 1,049/10,000 & 1,586 \\
		Square & 3,552/10,000 & 2,071 \\
		\bottomrule
	\end{tabular}
	\caption{Pearson-correlation-based detection on CIFAR test set with a coefficient $\rho \geq 0.5$.}
	\label{tab:PearsonCIFAR}
\end{table}

%
%


\textbf{Strategy 3: Zero effect}

In this strategy, we try to compute the number of samples having more than $n$ zero filters ($F_{Z_{0}}^{x\mapsto\varphi}$). We choose $n=100$  for our experiments on the CIFAR dataset. The results are reported in Table ~\ref{tab:ZeroCIFAR}. As we see from the results, natural or clean samples tend to have several zero filters, in contrary for all attacks, where the number of zero filters is decreased. This means that adversarial attacks try to exploit those non-causal features. 
While this strategy has been found effective with FGSM and BIM, for PGD and Square attacks, their results are not far enough from the clean samples with $7000$ and $5439$ samples having more than $100$ zero filters respectively. This proves again that both PGD and Square attacks are more resistant against our detection methods. Another reason for the samples not affected much by the strategy, is due to the low ASR, despite the high perturbation budget chosen $\epsilon=24$.

Generally, this strategy has been found promising only for CIFAR-10, and not for ImageNet, since ImageNet has a large number of classes (1k classes) with many complex features compared to CIFAR. So, the number of zero filters in ImageNet samples is very large for clean as well as adversarial samples in a similar way. 

%

\begin{table}[t]
	\renewcommand{\arraystretch}{1.0}
	\centering
	\begin{tabular}{lc}
		\toprule
		\textbf{Samples} & \textbf{Number ($N=100$)} \\
		\midrule
		Clean & 8,997 \\
		\midrule
		\multicolumn{2}{c}{\textit{Targeted}} \\
		\midrule
		FGSM & 1,455 \\
		PGD & 1,604 \\
		BIM & 1,040 \\
		\midrule
		\multicolumn{2}{c}{\textit{Untargeted}} \\
		\midrule
		FGSM & 2,041 \\
		PGD &  7,000\\
		BIM & 1,360 \\
		SQUARE & 5,439 \\
		\bottomrule
	\end{tabular}
	\caption{Number of CIFAR samples having higher than $N$ zero filters.}
	\label{tab:ZeroCIFAR}
\end{table}

\textbf{Strategy 4: Common Robust Causal Features:}

In this section, we aim to demonstrate that instances of the same prototype, if clean, exhibit several common top robust causal features. This stands in contrast to adversarial samples, which only share a few common features with their prototypes. The explanation for this is that the adversarial attacks try to exploit any possible activation towards another class despite the features in hand. In our experiments, we give several values for $n/m$: $3/10$, $5/20$ and $10/30$ respectively, all of them show the difference between clean and adversarial samples. We recall here that in this strategy, $n$ denotes the number of causal features in the input image (clean or adversarial), and $m$ denotes the number of causal features in the associated prototype image. The results on ImageNet dataset with $\epsilon=8$ are reported in Table  ~\ref{tab:TopCommon}. Across all configurations, we observe a clear distinction between clean and adversarial samples: clean instances consistently share more features with their prototypes (e.g. $1400$ and $1675$), whereas adversarial examples diverge significantly with few samples having common features (e.g. $13$ and $24$) for FGSM untargeted attack for instance.

Despite the parameters $n/m$, all the attacks produce similar results. When choosing 10/30 (10 for instance features and 30 for prototype features), the number of common causal features drops sharply, showing a significant difference. So as we see, there is a huge difference between clean images (1400 clean instances) and adversarial samples (only 16 samples for PGD untargeted attack) for example.

\begin{table}[t]
	\renewcommand{\arraystretch}{1.0}
	\centering
	\begin{tabular}{lccc}
		\toprule
		\textbf{Samples} & \textbf{Top (3/10)} & \textbf{Top (5/20)} & \textbf{Top (10/30)}\\
		\midrule
		Clean & 1,400 & 1,675 & 1,400 \\
		\midrule
		\multicolumn{4}{c}{\textit{Targeted}} \\
		\midrule
		FGSM & 59 & 133 & 29 \\
		PGD & 33 & 59 & 18   \\
		\midrule
		\multicolumn{4}{c}{\textit{Untargeted}} \\
		\midrule
		FGSM & 13 & 132 & 24\\
		PGD & 29 & 61 & 16\\
		Square & 737 & 969 & 704\\
		\bottomrule
	\end{tabular}
	\caption{Number of samples for clean and adversarial inputs on 6{,}000 ImageNet images under different $(n, m)$ top causal features.}
	
	\label{tab:TopCommon}
\end{table}


\subsection{Detection evaluation}

For evaluating the perofrmance of CausAdv in terms of detection metrics, we report in this section the performance metrics (TPR, Precision, and F1-score) of CausAdv for both Imagenet and CIFAR10. The results are reported in  Table~\ref{tab:causadv_imagenet_full} and Table~\ref{tab:causadv_cifar_full} respectively. These metrics are derived from the correlation-based results presented previously in Table~\ref{tab:PearsonImag} and Table~\ref{tab:PearsonCIFAR} .

The samples correlated are classified as clean, while others are considered adversarial. 
For ImageNet clean samples, the reported value (4,211 out of 6,000) corresponds to true negatives (TN), while the remaining samples are false positives (FP), yielding an FPR of 
\[
\text{FPR} = \frac{6000 - 4211}{6000} = 29.82\%.
\]
Since FPR is computed solely on clean samples, it remains constant across all attacks. 
For adversarial samples, the reported values correspond to false negatives (FN), and true positives (TP) are obtained as the complement with respect to the total number of adversarial samples. 
Finally, TPR, Precision, and F1-score are computed from the resulting confusion matrix.

\begin{table*}[t]
	\renewcommand{\arraystretch}{1.1}
	\centering
	\begin{tabular}{llc|cccc}
		\toprule
		\multirow{2}{*}{\textbf{Attack Type}} &
		\multirow{2}{*}{\textbf{Attack}} &
		\multirow{2}{*}{\textbf{ASR (\%)}} &
		\multicolumn{4}{c}{\textbf{CausAdv Performance (\%)}} \\
		\cmidrule(lr){4-7}
		& & & TPR & Precision & F1-score \\
		\midrule
		
		\multirow{4}{*}{Targeted}
		& FGSM & 91.38 & 91.43 & 75.42 & 82.65 \\
		& PGD  & 92.63 & 85.75 & 74.31 & 79.63 \\
		& BIM  & 93.15 & 100.00  & 77.05 & 87.04 \\
		& C\&W & 33.20 & 66.13  & 68.92 & 67.49 \\
		
		\midrule
		
		\multirow{5}{*}{Untargeted}
		& FGSM   & 93.55 & 81.68  & 73.26 & 77.26 \\
		& PGD    & 89.52 & 82.80  & 73.52 & 77.92 \\
		& BIM    & 96.40 & 100.00  & 77.05 & 87.04 \\
		& C\&W   & 36.23 & 68.30 & 69.63 & 68.96 \\
		& Square & 57.52 & 65.98 & 69.14 & 67.54 \\
		
		\bottomrule
	\end{tabular}
	
	\caption{Detection performance of CausAdv on ImageNet. 
	}
	\label{tab:causadv_imagenet_full}
\end{table*}

Overall, CausAdv achieves strong detection capability, with TPR values ranging from 65.98\% to 100\%. 
In particular, the highest detection rates are observed for the BIM attack, where both targeted and untargeted variants reach a perfect detection rate of 100\%, indicating a complete disruption of the underlying causal features captured by the model.

For other attacks such as FGSM and PGD, CausAdv maintains high detection rates, with TPR values exceeding 80\% in most cases, demonstrating robustness against strong gradient-based perturbations. 
However, lower detection rates are observed for attacks such as C\&W and Square, where TPR values decrease to approximately 66--68\%, suggesting that these attacks preserve partial correlation with class prototypes and are therefore more challenging to detect.

For CIFAR10, We report the detection performance metrics (TPR, Precision, and F1-score) of CausAdv in Table~\ref{tab:causadv_cifar_full}, with an FPR of \[
\text{FPR} = \frac{10000 - 8986}{10000} = 10.14\%.
\].

CausAdv achieves a significantly lower false positive rate (FPR = 10.14\%) on CIFAR-10 compared to ImageNet (FPR = 29.82\%). 
This improvement can be attributed to the smaller number of classes (10 vs. 1000), along with reduced intra-class variability and lower visual complexity in CIFAR-10, leading to more stable and separable correlation between clean and adversarial samples.

Overall, the method achieves consistently strong detection performance, with TPR values ranging from  84.56 \% to 88.43\% for targeted attacks and from  69.74\% to 87.36\% for untargeted attacks. 
In particular, the highest detection rates are observed for BIM and FGSM attacks, indicating that these perturbations significantly disrupt the correlation with class prototypes.

In addition to high TPR, CausAdv maintains strong precision across all attack scenarios, with values consistently above 87\%, reflecting a low number of false positives. 


Overall, these results indicate that CausAdv provides reliable and stable detection performance on CIFAR-10, achieving an effective trade-off between sensitivity to adversarial perturbations and robustness on clean samples.

\begin{table*}[t]
	\renewcommand{\arraystretch}{1.1}
	\centering
	\begin{tabular}{llc|cccc}
		\toprule
		\multirow{2}{*}{\textbf{Attack Type}} &
		\multirow{2}{*}{\textbf{Attack}} &
		\multirow{2}{*}{\textbf{ASR (\%)}} &
		\multicolumn{4}{c}{\textbf{CausAdv Performance (\%)}} \\
		\cmidrule(lr){4-7}
		& & & TPR & Precision & F1-score \\
		\midrule
		
		\multirow{3}{*}{Targeted}
		& FGSM & 9.83 & 84.56 & 89.29 & 86.86 \\
		& PGD  & 7.09 & 86.28 & 89.48 & 87.84 \\
		& BIM  & 6.09 & 88.43  & 89.71 & 89.06 \\
		
		\midrule
		
		\multirow{4}{*}{Untargeted}
		& FGSM   & 66.18 & 87.36  & 89.63 & 88.47 \\
		& PGD    & 10.49 & 69.74  & 87.30 & 77.53 \\
		& BIM    & 10.49 & 84.14 & 89.19 & 86.59 \\
		& Square & 35.52 & 79.29  & 88.67 & 83.70 \\
		
		\bottomrule
	\end{tabular}
	
	\caption{Detection performance of CausAdv on CIFAR-10 at threshold $\rho \geq 0.5$. 
	}
	\label{tab:causadv_cifar_full}
\end{table*}

\subsection{Comparison with existing detection methods}
Table~\ref{tab:Defense} presents the results of CausAdv compared to four defense methods: Feature Squeezing (FS)\cite{Weilin_Squeez}, Spatial Smoothing (SS) \cite{Weilin_Squeez}, Gaussian Data Augmentation (GDA) \cite{GDA} and CutMix \cite{SangdooCutMix}, evaluated on the same set of $6,000$ adversarial samples from the ImageNet dataset. We employ the True Positive Rate (TPR) or Recall as a measure of performance on the adversarial examples.
All defense methods were implemented using the Adversarial Robustness Toolbox (ART) library \cite{ART}, under consistent experimental settings. As shown in the results, for a perturbation budget of $\epsilon = 8$, most defense methods fail to recover the correct classes for the adversarial samples, whether the attacks are targeted or untargeted, especially for Square and C\&W, which is due mainly to the low ASR of these attacks in the first place compared to other attacks as reported in Table ~\ref{tab:PearsonImag}, in addition to their efficiency. By comparing defense methods, Spatial Smoothing (SS) performs slightly better than the other methods, retrieving a higher number of correct predictions, specifically for PGD and BIM attacks but its effectiveness remains limited overall.

These findings highlight the difficulty of defending against adversarial examples on large-scale datasets like ImageNet. In contrast, our proposed method, CausAdv, significantly outperforms all tested defenses, particularly in handling BIM and FGSM-targeted attacks. CausAdv achieves a lower detection rate only for the Square and C\&W attacks because these attacks exhibit a lower ASR compared to other attacks. Consequently, the perturbations introduced by Square and C\&W are relatively weak, causing the adversarial samples to appear similar to clean samples. As a result, the causal features are not significantly altered, making them harder for CausAdv to distinguish from normal samples.

\begin{table*}[t]
	\renewcommand{\arraystretch}{1.1}
	\centering
	\begin{tabular}{llc|ccccc}
		\toprule
		\multirow{2}{*}{\textbf{Attack Type}} &
		\multirow{2}{*}{\textbf{Attack}} &
		\multirow{2}{*}{\textbf{ASR (\%)}} &
		\multicolumn{5}{c}{\textbf{TPR (\%) by Defense Method}} \\
		\cmidrule(lr){4-8}
		& & & FS & SS & GDA & CutMix & \textbf{CausAdv} \\
		\midrule
		
		\multirow{3}{*}{Targeted}
		& FGSM & 91.38 & 8.40 & 8.58 & 8.58 & 8.61 & \textbf{91.43} \\
		& PGD  & 92.63 & 12.15 & 31.33 & 7.56 & 7.36 & \textbf{85.75} \\
		& BIM  & 93.15 & 11.16 & 30.21 & 6.81 & 6.85 & \textbf{100} \\
		& C\&W & 33.20 & 10.76 & 12.51 & 3.03 & 0.0 & \textbf{66.13} \\
		\midrule
		
		\multirow{4}{*}{Untargeted}
		& FGSM   & 93.55 & 6.73 & 10.90 & 6.41 & 6.45 & \textbf{81.68} \\
		& PGD    & 89.52 & 16.86 & 27.16 & 16.86 & 10.48 & \textbf{82.80} \\
		& BIM    & 96.40 & 2.68 & 4.38 & 2.58 & 2.58 & \textbf{100} \\
		& C\&W & 36.23 & 7.45 & 9.01 & 3.08 & 0.0 & \textbf{68.30} \\
		& Square & 57.52 & 10.11 & 15.73 & 4.34 & 0.0 & \textbf{65.98} \\
		\bottomrule
	\end{tabular}
	
	\caption{True Positive Rate(TPR) of each defense method for both targeted and untargeted attacks ($\epsilon=8$) given 6,000 imagenet samples.}
	\label{tab:Defense}
\end{table*}

\subsection{Similarity to other works and discussion}
Several researchers have observed a fundamental distinction between natural images and artificially perturbed ones. For instance, \cite{Tao_Turning} propose multiple criteria for determining whether an image is benign, all based on two key assumptions: (1) natural images should yield high prediction confidence near the decision boundary, and (2) natural images are expected to be robust to random noise \cite{Szegedy}. However, evaluating these criteria requires injecting noise into the input images to test robustness under perturbation.

Another work \cite{Ilyas_Bugs} introduces a different assumption: classifiers are trained to solely maximize accuracy, for reaching this goal, they tend to use any available activations, even those that refer to features that could look incomprehensible to humans.  Interestingly, rather than blaming on the models, they attribute the vulnerability to the input images themselves, claiming these non-robust features are intrinsic to the data and can be easily exploited through adversarial perturbations. Consequently, they propose modifying input images by identifying and removing such non-robust features.

Both of the the mentioned works aim to modify input images, either by introducing noise or by removing non-robust features. In contrast, our method does not require any modification to the input image. While our approach shares conceptual similarities with \cite{Ilyas_Bugs} in distinguishing between robust and non-robust features, it differs in execution. Specifically, we identify causal filters, those with positive Causal Importance (CI) values, which correspond to robust features. Furthermore, our method captures a gradual robustness: filters with higher CI values are considered more robust than those with lower but still positive CI values. Filters with zero or negative CI values are interpreted as non-causal, aligning with what \cite{Ilyas_Bugs} refer to as non-robust features. Our experimental results on the CIFAR-10 dataset further support this interpretation. As shown in Table~\ref{tab:ZeroCIFAR}, adversarial attacks tend to exploit non-robust features, those associated with CI values of zero, confirming the connection between adversarial vulnerability and non-causal activations.

Our approach also shares a conceptual similarity with \cite{Tao_Turning} in how CI values are computed. Specifically, the counterfactual information (CI) of each filter is derived based on changes in the model's prediction probability, under the same underlying assumption that the absence of robust features leads to a significant drop in prediction confidence. This supports the broader idea that natural images should be classified with high confidence when near the decision boundary. This trend is especially evident in the case of FGSM non-targeted attacks, where all perturbed images exhibit notably low prediction probabilities.

However, our methodology differs in how robustness is quantified. While the referenced work defines the decision boundary in terms of the number of gradient steps required to alter a prediction, we instead measure robustness through CI values, by directly observing the change in prediction probability when individual filters are intervened upon. This provides a more granular, feature-level view of robustness that does not require iterative modification of the input.


In terms of statistical analysis, our work has many similarities with \cite{Roth_Stat,Xin_Stat,Grosse,Feinman}. Roth et al. \cite{Roth_Stat} tend to measure how features and log-odds change under noise, assuming that: if inputs are adversarially perturbed, the variation in noise-induced features may tend to have a characteristic direction, in contrast, with natural input images, it should not have any
specific direction. Our approach, which shares similarities with this work, can even be visually captured statistically. This is achieved by plotting the distribution of CI values for each input image, revealing a significant variation with adversarial inputs, particularly with FGSM attacks. 
While their work measures expected log-odds of gradients at test-time and adding noise as well, CausAdv performs causal learning at test time as well, without any need to alter the input image.

Overall, our causal approach yields promising results for almost every attack for the two datasets ImageNet and CIFAR-10. For FGSM, it provided good results for both datasets, especially the target variant on ImageNet (See Table~\ref{tab:PearsonImag}). For BIM, although it is a very efficient attack \cite{Dong_Benchmark}, we find that it is easily detected by identifying the set of causal features on ImageNet. Only PGD exhibit a clear resistance to our defense strategies under the chosen perturbation budget ($\epsilon = 8$). However, increasing the perturbation strength to $\epsilon = 16$ leads to complete detection for PGD, similar to the BIM attack. These findings are consistent with those reported in \cite{Dong_Benchmark}, which highlight PGD as the most suitable attack for adversarial training. For C\&W and Square attacks, they exhibit higher resistance, but with low Attack Success Rate(ASR).


In conclusion, our analysis aligns with the findings reported in \cite{Ilyas_Bugs}, which argue that adversarial vulnerabilities do not stem from flaws in the model itself, but rather from its reliance on non-robust features. However, while their work emphasizes this insight, it does not explicitly demonstrate how non-robust features can be identified or linked to adversarial perturbations. Our framework addresses this gap by providing a practical means to detect such features. 


\section{Detection through the interpretability of causal features}

Beyond adversarial attacks, deep learning models continue to face a fundamental challenge: the interpretability and explainability of their decisions. In many applications, trustworthiness of these models is critical. As a result, the field of explainable Artificial Intelligence (XAI) has emerged to increase our confidence in these models. However, \cite{BANIECKI2024102303} highlights a new  cross-domain research area called adversarial explainable AI (AdvXAI), by showing that explainability methods themselves can be vulnerable to adversarial attacks. Furthermore, \cite{CHOU202259} investigates the role of counterfactual reasoning in explanation and argues that most existing XAI methods often reflect spurious correlations rather than true causal relationships. The authors claim that only counterfactual explanations offer a path toward genuine causality, which in turn enables robust and faithful interpretations. In this section, we will show the strong relationship between robustness and explainability by exploring the additional useful property of CausAdv, which is interpretability.

Among the most used explanation techniques are gradient-based methods including Class Activation Mapping (CAM) \cite{Zhou07}, Grad-CAM \cite{Selvaraju08}, and SmoothGrad \cite{Simonyan10}. These approaches aim to identify important neural activations and generate attention maps by repeatedly probing the network to highlight the most discriminative regions in input images. CausAdv, shares conceptual similarities with CAM and Grad-CAM, as it is score-based approach for identifying important filters. However, CausAdv distinguishes itself by providing a more robust and causally grounded formulation. Rather than relying on gradients with respect to the input instance, CausAdv computes the causal contribution CI scores using counterfactual reasoning, evaluating the impact of individual filters on the prediction probability of a prototype class. These CI scores can then be used to localize the most discriminative regions of any input, offering an interpretable and more stable alternative to traditional gradient-based methods. In contrast, Grad-CAM generates scores based on gradients with respect to the current input, making it more susceptible to input-specific noise and adversarial perturbations.

\begin{figure}[h!]
	\centering
	\includegraphics[width=\linewidth]{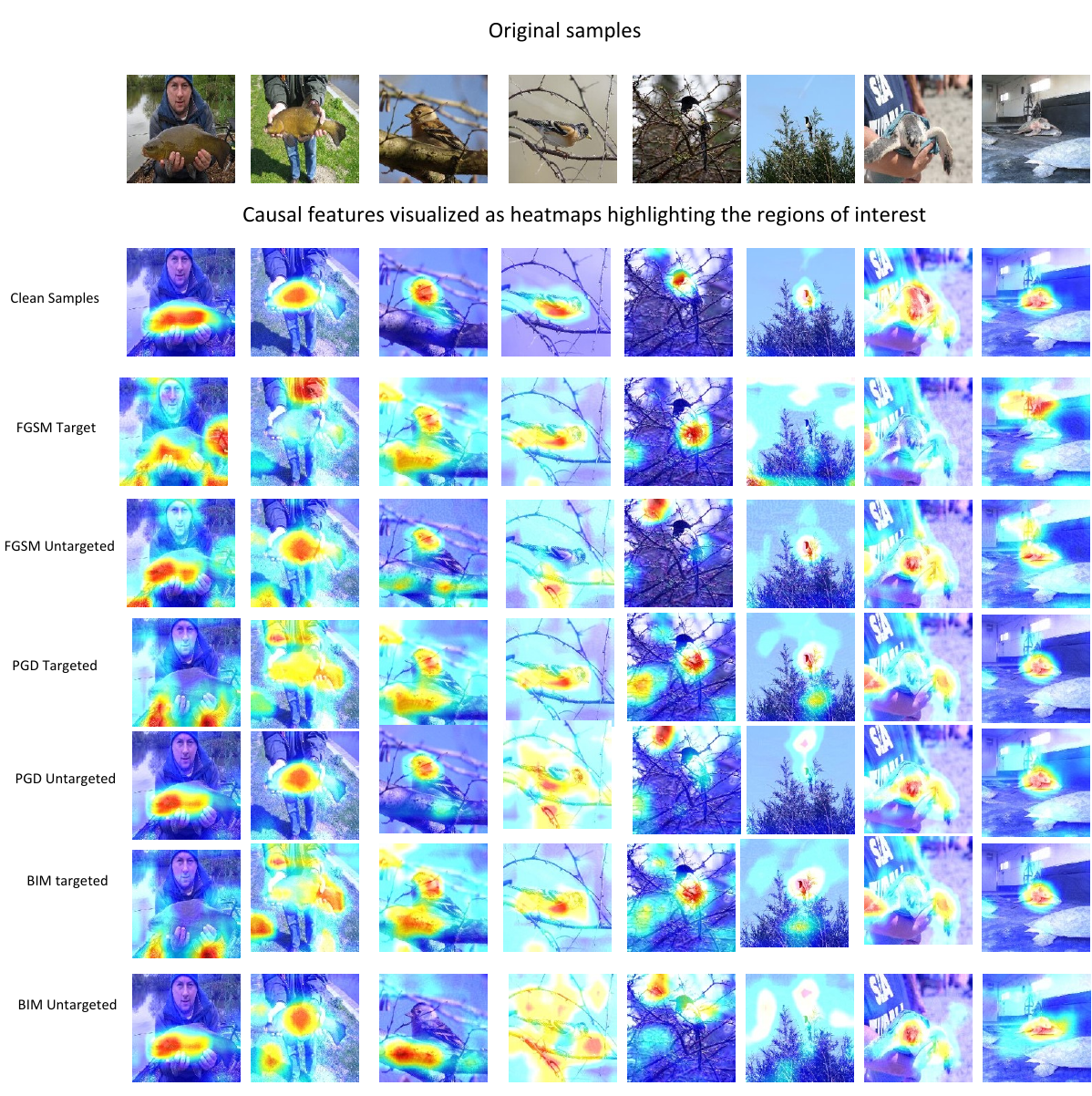}
	\caption{Causal features visualized as attention maps on clean and adversarial samples.}
	\label{fig:Explain}
\end{figure}


Since the explanations provided by our framework are very robust compared to gradient-based techniques, we suggest to use them as a qualitative adversarial detection technique as follow: For every prototype class $x_{\varphi}^{prot}$, we identify the most discriminative features based on the filters' CI from most important to least important, where only causal filters $F_{W}^{x_{\varphi}^{prot}\mapsto\varphi}$ are counted. These features can be then visualized as attention maps and serve as reference for every instance of this prototype. To determine whether a sample $x$ (predicted as class $\varphi$) is clean or adversarial, we overlay the causal features of the prototype $x_{\varphi}^{\text{prot}}$ onto the sample $x$. If $x$ is clean, we expect the visualized features to align well with those of the prototype, maintaining semantic and spatial coherence. In contrast, if $x$ is adversarial, the visualized causal features on $x$ are likely to appear significantly different, semantically irrelevant, or lacking interpretability, thus serving as a strong qualitative indicator of adversarial behavior.

Figure~\ref{fig:Explain} presents qualitative results for our adversarial detection approach based on causal explanations. The figure includes original input samples along with heatmaps highlighting their most discriminative causal features, visualized as attention maps. 
As shown, for clean samples, the heatmaps generated using the causal features associated with the true class offer clear and coherent visual explanations of the model's decisions. These maps consistently highlight the relevant regions of interest that contribute to the prediction and are generally stable across different instances of the same class. This consistency reinforces the idea that the features responsible for the decision are truly present in the image, thus offering strong evidence of the prediction's reliability. In contrast, for adversarial examples across all attack types, the generated heatmaps lack interpretability and appear semantically meaningless. This is because the causal features of the adversarially predicted class are not genuinely present in the input image. As a result, the corresponding attention maps fail to highlight any meaningful regions, making them ineffective as explanations. These observations clearly demonstrate the complementary role of causal explanation in enhancing adversarial robustness. These findings highlight the potential of this method for real-world applications such as object localization and detection.

\section{Conclusion and Future Work}

In this paper, we introduced CausAdv, a causal framework for detecting adversarial examples based on counterfactual reasoning. CausAdv leverages statistical analysis of Counterfactual Information (CI) values across convolutional filters to identify anomalies in their distribution, enabling reliable distinction between natural and adversarial inputs.

Extensive experiments demonstrate that CausAdv effectively detect adversarial examples across various attacks, achieving 100\% detection accuracy against BIM  on the ImageNet, significantly outperforming existing defense methods. The framework also supports multiple detection strategies, ensuring robustness against a range of attack behaviors: if one strategy underperforms for a particular attack, another may succeed. Moreover, the causal features identified by CausAdv offer interpretable visual explanations, enabling clear distinction between clean and adversarial samples. These findings highlight the potential of causal reasoning as a powerful, and explainable approach to improve adversarial robustness in CNNs.

\bibliographystyle{elsarticle-num-names} 
\bibliography{AdvAttBib}

\end{document}